\pdfoutput=1
\documentclass[11pt]{article}

\usepackage{acl}
\usepackage{hyperref}
\usepackage{times}
\usepackage{latexsym}
\usepackage{multirow}
\usepackage{booktabs}
\usepackage{microtype}
\usepackage{multirow}
\usepackage{latexsym}
\usepackage{booktabs}
\usepackage{courier}
\usepackage{amsmath}
\usepackage{booktabs} % For better table formatting
\usepackage{siunitx} % For aligning numbers by decimal point
\usepackage{subfig}
\usepackage{placeins}
\usepackage{tikz}
\usepackage{hyperref}
\usepackage[symbol]{footmisc}
\usepackage{adjustbox}
\usepackage{xurl}

\usepackage[T1]{fontenc}
\usepackage[utf8]{inputenc}

\usepackage{microtype}

\usepackage{inconsolata}

\title{TLDR at SemEval-2024 Task 2: T5-generated clinical-Language summaries for DeBERTa Report Analysis}

\author{Spandan Das \thanks{All authors contributed equally to this work.} \\
  Carnegie Mellon University \\
  \texttt{spandand@andrew.cmu.edu} \\\And
  Vinay Samuel \footnotemark[1] \\
  Carnegie Mellon University \\
  \texttt{vsamuel@andrew.cmu.edu} \\\And
   Shahriar Noroozizadeh \footnotemark[1] \\
  Machine Learning Department \\
  Carnegie Mellon University \\
  \texttt{snoroozi@cs.cmu.edu}}
\begin{document}
\maketitle
\begin{abstract}
    This paper introduces novel methodologies for the Natural Language Inference for Clinical Trials (NLI4CT) task. We present TLDR (\textbf{T}5-generated clinical-\textbf{L}anguage summaries for \textbf{D}eBERTa \textbf{R}eport Analysis) which incorporates T5-model generated premise summaries for improved entailment and contradiction analysis in clinical NLI tasks. This approach overcomes the challenges posed by small context windows and lengthy premises, leading to a substantial improvement in Macro F1 scores: a 0.184 increase over truncated premises. Our comprehensive experimental evaluation, including detailed error analysis and ablations, confirms the superiority of TLDR in achieving consistency and faithfulness in predictions against semantically altered inputs.
\end{abstract}

\section{Introduction}
The Multi-evidence Natural Language Inference for Clinical Trial Data (NLI4CT) task focuses on developing systems that can interpret clinical trial reports (CTRs) and make inferences about them \citet{jullien-etal-2024-semeval}. The task provides a collection of breast cancer CTRs from ClinicalTrials.gov along with hypothesis statements and labels annotated by clinical experts.

The NLI4CT 2024 task is to classify whether a given CTR entails or contradicts the hypothesis statement. This is challenging as it requires aggregating heterogeneous evidence from different sections of the CTRs like interventions, results, and adverse events. The dataset contains 999 breast cancer CTRs and 2400 annotated hypothesis statements split into training, development, and test sets. The CTRs are summarized into sections aligning with Patient Intervention Comparison Outcome framework. The 2024 SemEval task has the same training dataset as the SemEval 2023 task \citep{jullien-etal-2023-semeval} but the development and test sets for the 2024 task include interventions of either preserving or inversing the entailment relations for some data points. 

In this paper, we introduce TLDR (\textbf{T}5-generated clinical-\textbf{L}anguage summaries for \textbf{D}eBERTa \textbf{R}eport Analysis), a novel framework developed for the SemEval Task 2024. Our key contribution is the integration of a T5-based summarization approach to preprocess and condense the premises of clinical reports, which are then analyzed alongside the corresponding statements using DeBERTa, an encoder-only transformer to perform Natural Language Inference (NLI). Our T5-generated summaries address the limitations of small context windows and lengthy premises for this task. This methodology demonstrates a significant improvement in performance on the held-out test set, with our best model achieving an increase of 0.184 in the Macro F1 score compared to using truncated premises and 0.046 over extractive summarized premises. To underscore the efficacy of our approach in this task, we have conducted extensive experiments and ablations, complemented by a thorough error analysis. We also demonstrate the efficacy of our model's performance with regards to consistency and faithfulness against semantically altered inputs. These efforts collectively highlight the robustness and effectiveness of the TLDR framework in addressing the complexities and nuances of NLI task within the domain of clinical report trials.

\begin{figure*}[ht!]
\centering
\includegraphics[width=\linewidth]{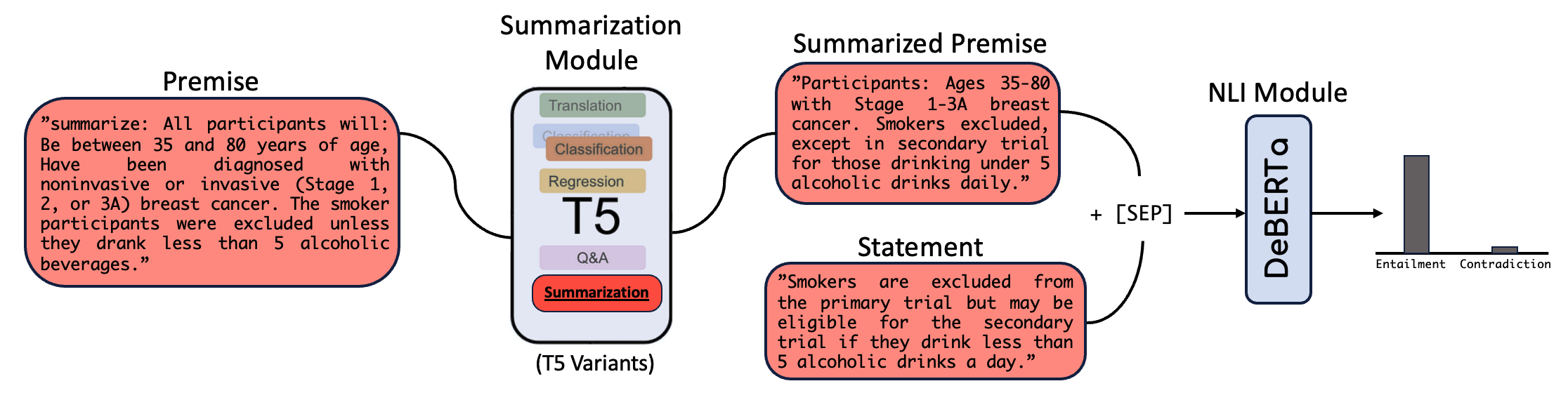}
\caption{TLDR Model: TLDR processes the clinical report by initially summarizing it using the summarization module (variants of the T5 Model). This summary is then merged with the statement and fed into the fine-tuned DeBERTa model for Natural Language Inference.}
\label{fig:summary}
\end{figure*}

\section{Background}
% \vspace{-0.5em}
\label{sec:background}
The NLI4CT task requires developing systems capable of NLI from clinical trial reports. In the following section, we examine not only contributions within the SemEval NLI4CT task \citep{jullien-etal-2023-semeval, jullien-etal-2023-nli4ct} but also wider advancements in the field. 
% \vspace{-.3em}
\paragraph{Transformer Architectures}
Pretrained transformer models form the backbone of many top-performing systems in the NLI4CT task, with their architectures being a crucial aspect of their design.

\subparagraph{Generative Transformers}
Generative transformers, which include models like the instruction-tuned Flan-T5 \citep{kanakarajan-sankarasubbu-2023-saama}, are particularly adept at generating text and can produce probabilities or direct entailment labels. These models excel in tasks that require the generation of coherent language constructs and have been pretrained on biomedical data, equipping them with the necessary domain knowledge. The work of \citet{zhao2023hw} further emphasizes this, showcasing the efficacy of ChatGPT, a generative model, in a multi-strategy system for clinical trial inference, particularly through prompt learning techniques.

\subparagraph{Discriminative Transformers}
Discriminative transformers are employed for classification tasks and include variants of BERT such as BioBERT and ClinicalBERT \citep{lee2020biobert, chen-etal-2023-ncuee-nlp, vladika-matthes-2023-sebis}. These models have been fine-tuned on domain-specific data to enhance their understanding of medical texts. DeBERTa architecture \cite{he2020deberta}, which improves upon BERT with disentangled attention and enhanced masking, is also included in this category. These approaches were successfully incorporated by \citet{chen-etal-2023-ncuee-nlp} and \citet{alameldin2023clemson} utilizing transformer-based models for both evidence retrieval and entailment determination in NLI4CT task of 2023.

\section{System overview}
% \vspace{-0.6em}
In this section, we explain our methodologies to address the complexities of the Natural Language Inference for Clinical Trials (NLI4CT) task. We primarily focus on the utilization of large language models such as T5 \citep{raffel2020exploring} and BERT-based architectures \citep{devlin2018bert}, to implement innovative techniques to manage the challenge of the long premise lengths and token size limitations.
As mentioned in Section~\ref{sec:background}, DeBERTa is a top-performing model for NLI tasks. However, it has a limitation in clinical NLI such as the task in NLI4CT due to its restricted context length, making it difficult to include both the premise and the statement. We therefore propose to fine-tune T5 to summarize long clinical premises, which can then allow us to leverage DeBERTa's discriminative power for clinical NLI with extended premises.
We call our model TLDR for \textbf{T}5-generated clinical-\textbf{L}anguage summaries for \textbf{D}eBERTa \textbf{R}eport Analysis. 
Our full pipeline can be seen in Figure \ref{fig:summary}.

In this section, we delve into the fine-tuning of these models for our clinical NLI task and employ specialized summarization strategies using variants of the T5 model, balancing between zero-shot and fine-tuned approaches. 
\footnote{Our code is available at: \\ \url{https://github.com/Shahriarnz14/TLDR-T5-generated-clinical-Language-for-DeBERTa-Report-Analysis}}

\subsection{TLDR: NLI with DeBERTa enhanced by T5 Generated Summaries}
% \vspace{-.3em}
We experimented with DeBERTa, an encoder-only transformer model and its ability for natural language inference for the NLI4CT task. Taking the context length into account, the full premise does not fit into our model so we would have to get a shortened version of the premise. To achieve this, we generate a summary of each premise using a T5 variant that can fit the context length of DeBERTa (see Section~\ref{sec:summarization}). This shortened premise along with the full statement of each data point in the dataset is then used to make a prediction of entailment or contradiction for that statement. Our input to the DeBERTa model has the following form:
\begin{center}
  \textbf{\textit{[CLS] + shortened premise + [SEP] + statement}}
\end{center} where the [CLS] token is used for the binary classification of entailment or contradiction.

\subsection{Summarization Techniques for Premises}
\label{sec:summarization}
The need for summarizing the premises arises due to the long length of the premises and the limited input length of several of the top performing BERT-based architectures such as DeBERTa that is utilized in this paper.

In our tasks, there are two types of data points: "Single" and "Comparison". For "Single" data points, the statement is checked for entailment with only a primary premise. In "Comparison" instances the statement is checked for entailment when compared to a primary as well as a secondary premise. For each of the summarization methods below, we consider summarizing both the primary and second premises independently of each other as well as summarizing the primary and secondary premises combined together.

\subsubsection{Inference Only}
% \vspace{-.3em}
Encoder-Decoder architectures such as the T5 model have shown strong capabilities for summarizing text. To this end we used several T5 variants to produce a summary for each premise with a maximum source length of 2048 tokens and a maximum generated summary of 300 tokens thereby decreasing the size of the premise to a size that would enable our full input to be passed into the DeBERTa model.

\subsubsection{Fine-tuning}
% \vspace{-.3em}
Our objective for fine-tuning T5-based models is to enable the generation of summaries closely mirroring ground truth statements for entailed premise-statement pairs. We fine-tune T5 variants on a dataset exclusively comprising pairs labeled "Entailment". "Contradiction" instances are excluded to avoid confusion of the model in generating summaries as they include contradictory information about the premise. We prepend the premises with "summarize:", as inputs to T5 and treat the corresponding ground truth statements as labels, aiming to align the generated summaries with these statements. This approach ensures the model is trained to produce summaries that effectively encapsulate the entailed information.

\section{Experimental setup}
% \vspace{-0.6em}
In this section, we provide a comprehensive outline of our experimental methods, setting the stage for a detailed discussion on each technique employed. 

The fine-tuning of each module of our model was done on the training set and evaluated on the development set. Upon selecting the best performing model from the development set, we then evaluated our model on the held-out test. The performance reported in Section~\ref{sec:results} is on this held-out test set. 

\subsection{The Discriminative NLI Module of TLDR}
% \vspace{-.3em}
For the shortened premise and the statement pairs, we fine-tuned an NLI fine-tuned version of the base DeBERTa-v3 model \citep{tran-etal-2023-videberta} from Hugging Face for Entailment and Contradiction binary classification. Specifically we used the \url{cross-encoder/nli-deberta-v3-base}. This model was trained using SentenceTransformers Cross-Encoder class on the SNLI \citep{bowman2015large} and MultiNLI \citep{N18-1101} datasets and provided improved NLI performance over standard DeBERTa-v3-base model. The tokenizer used was also taken from Hugging Face and was the tokenizer corresponding to the DeBERTa model that we used. The fine-tuning of our DeBERTa model was done on the training set for 40 epochs at a learning rate of $5 \times 10^{-5} $ and evaluated on the development set to pick the best performing model for the final evaluations.

Importantly, DeBERTa has a maximum input size of 512 tokens. Therefore, the combined length of the shortened premise and statement pair is required to fit into this size (the average statement length in the training data was 110 tokens). The description of the different premise shortening techniques we employed is outlined below.

\subsection{The Summarization Module of TLDR}
% \vspace{-.3em}
We employed different variants of T5 to generate summaries of the clinical premises as a way to shorten them. In this section, we explain each of these approaches.
\paragraph{Zero-Shot T5}
We utilized the Hugging Face API to summarize premises with \textit{google/flan-t5-base}, limiting summaries to 300 tokens. For instances containing both primary and secondary premises, we conducted two different approaches: (1) separate summaries for each and (2) a combined summary for both premises together. The first approach involved generating individual summaries for primary and secondary separately, and creating a single summary for a concatenated version of both premises in the second approach.

\paragraph{Fine-tuned T5 For Summary Generation}
To fine-tune T5 for summary generation, we filtered our training and development datasets to only include instances labeled as entailment, using these as ground truth for summarized premises. In the case of entailment, we claim that since the statement is an accurate representation of the premise it also serves as an appropriate ground truth label for a summarized premise. We opted for the \textit{t5-small} model due to resource constraints, acknowledging this may impact comparison fairness with zero-shot models. Our fine-tuning utilized the ROUGE-1 metric which compares unigrams between the predicted and the label summary. We fine-tuned for $2, 5, 7, \text{and } 10$ epochs at a learning rate of $2 \times 10^{-5}$, weight decay of $0.01$ and a batch size of $4$.After fine-tuning, we generated separate summaries for primary and secondary premises using this model.

\paragraph{Fine-tuned SciFive}
We followed the exact same procedure explained above from the fine-tuned T5 for the \textit{razent/SciFive-base-Pubmed\_PMC} model from Hugging Face. For this T5 variant, we aimed leverage the fact that the SciFive model was trained on biomedical literature \citep{phan2021scifive} which is similar in domain to our setting. Here, we used the same hyperparameters for fine-tuning T5. After fine-tuning was complete, for data instances with two premises, we generated the summaries by separately summarizing the primary and secondary premises and then combining the two.

\subsection{Summarization Ablations}
We used two ablation strategies for TLDR's summarization module. Instead of using a T5 variant for summarizing each premise, these two ablations included: (1) naively truncating premises and (2) using a traditional extractive summarization.

\paragraph{Truncated Premises}
To fit within the 512 token limit of DeBERTa, we tokenized the combined statement and premises, subtracting the statement's token count and an additional 10 tokens from 512 to stay under the limit. 
More concretely, for 
$x = 512 - \text{(\# of statement tokens)} - 10$, 
in single-premise data points, we truncated the primary premise to $x$ tokens and in comparison data points with both a primary and a secondary premise, we truncated each premise to $\frac{x}{2}$ tokens.

\paragraph{Extractive Summarization}
We also used an extractive summarization technique with TF-IDF for our ablation experiments, which evaluates word significance in a document against a corpus. To avoid data leakage and maintain evaluation accuracy, the TF-IDF vectorizer was applied exclusively to the training dataset. It was then used to summarize texts within the training, development, and test sets. Summaries were produced by identifying and selecting sentences with the highest TF-IDF scores, adhering to a 300-word limit for conciseness. This extractive summarization ablation, based on TF-IDF, allows us to compare summarization techniques and their impact on DeBERTa's NLI model performance, emphasizing the importance of feature representation and data handling.

\subsection{Performance Metrics}
We report the held-out test set performance metrics of the different variants of TLDR model. Specifically, we are report Macro F1, Precision, and Recall as measures of the prediction performance. In addition, we also report Faithfulness and Consistency.

\textbf{Faithfulness} assesses if a system predicts accurately for the right reasons, gauged by its response to \textbf{semantic altering} interventions. With $N$ contrast set statements $x_i$, original $y_i$, and predictions $f()$, faithfulness is calculated using Equation~\ref{eq:faithfulness}.
%\vspace{-0.6em}
\begin{equation}
\footnotesize
    \begin{aligned}
   Faithfulness = \frac{1}{N}\sum_{1}^{N}\left| f(y_i)-f(x_i) \right|\hspace{1cm}\\ x_i\in C:\text{Label}(x_i) \neq \text{Label}(y_i), \text{ and } f(y_i) = \text{Label}(y_i) 
    \end{aligned}
    \label{eq:faithfulness}
\end{equation}

\textbf{Consistency} evaluates a system's output uniformity for semantically equivalent inputs, focusing on identical predictions under \textbf{semantic preserving} interventions. This ensures semantic representation consistency, regardless of prediction accuracy. For $N$ statements $x_i$ in contrast set ($C$), with original $y_i$ and predictions $f()$, consistency is determined using Equation~\ref{eq:consistency}.
%\vspace{-0.6em}
\begin{equation}
\footnotesize
    \begin{aligned}
   Consistency = \frac{1}{N}\sum_{1}^{N} 1 - \left| f(y_i)-f(x_i) \right| \\ x_i\in C:\text{Label}(x_i) = \text{Label}(y_i)
    \end{aligned}
    \label{eq:consistency}
\end{equation}

\begin{table*}[!ht]
\centering
\caption{Performance of TLDR Variants and Ablations on the Held-Out Test Set}
% \vspace{-.9em}}
\label{tab:main_results}
\adjustbox{max width=1\linewidth}{
\begin{tabular}{|l||c|c|c|c|c|}
\hline
% \rowcolor{gray!30}
\textbf{Method} & \textbf{Macro F1} & \textbf{Precision} & \textbf{Recall} & \textbf{Faithfulness} & \textbf{Consistency} \\ \hline \hline
% \rowcolor{gray!15}
\multicolumn{6}{|c|}{\textbf{Ablations: DeBERTa Methods}} \\ \hline
DeBERTa + Truncated Premise(s) &  0.474 & 0.432 & 0.524 & 0.573 & 0.542 \\ \hline
DeBERTa + Extractive Summarized Premise(s) &  0.612 & 0.584 & \textbf{0.643} & \textbf{0.615} & \textbf{0.590} \\ \hline
% \rowcolor{gray!15}
\multicolumn{6}{|c|}{\textbf{TLDR Methods}} \\ \hline
TLDR (flan-T5-base - Zero-Shot Combined Premises) &  0.599 & 0.628 & 0.573 & 0.409 & 0.540 \\ \hline
TLDR (flan-T5-base - Zero-Shot Distinct Premises) &  0.633 & 0.624 & 0.642 & 0.502 & 0.574 \\ \hline
TLDR (T5-small - Best fine-tuned) &  0.635 & 0.676 & 0.599 & 0.436 & 0.557 \\ \hline
TLDR (SciFive-base - Best fine-tuned) &  \textbf{0.658} & \textbf{0.684} & 0.633 & 0.501 & 0.581 \\ \hline
\end{tabular}}
\end{table*}
% \vspace{-1em}
\section{Results}
\label{sec:results}
% \vspace{-0.6em}
The performance of our TLDR variants and ablations on the NLI task for clinical report trials is summarized in Table~\ref{tab:main_results}. Our results showcase the effectiveness of different approaches, with notable variations in Macro F1, Precision, Recall, and Faithfulness, and Consistency across methods.

The TLDR methods showed the most promising results. The best-performing model based on prediction performance was the TLDR method using SciFive-base for fine-tuned summarization of distinct premises, achieving the highest Macro F1 score of 0.658 and the best precision of 0.684. This approach also demonstrated a strong recall at 0.633 thereby indicating a strong balance between precision and recall. A close second was the TLDR approach with fine-tuned T5-small, which attained a Macro F1 score of 0.635 and precision of 0.676 and recall of 0.599, indicating the efficacy of fine-tuning on task-specific data. In the appendix, Table~\ref{tab:finetuning} illustrates the impact of varying the total number of fine-tuning epochs for T5-small and SciFive on the downstream NLI task. The main finding is that unlike T5-small that longer fine-tuning degraded the downstream performance, SciFive required longer fine-tuning steps to see improvements in TLDR's downstream prediction. We suspect this is due to the fact that the original fine-tuning of SciFive had degraded its summarization performance compared to T5-small and thus it required to be fine-tuned for longer.

For the ablations methods, we observed an improvement in performance when using extractive summarization instead of naively truncating the input. The method utilizing truncated premises achieved a Macro F1 score of 0.474, with the lowest precision of 0.432 among all methods and the lowest recall of 0.524 among all methods, suggesting this strategy as being inappropriate for dealing with the context length issue for this particular task. The extractive summarization approach yielded a much higher Macro F1 score of 0.612 with a much higher precision at 0.584 and recall at 0.643 thereby clearly outperforming the truncated premises strategy. Note that these are ablations of our introduced more complete TLDR model. Extractive summarization proved the best strategy for the faithfulness and consistency metrics with a faithfulness score of 0.615 and a consistency of 0.590.
This result likely stems from the fact that extractive summarization of premises maintains key tokens from the original premises, which preserves the core semantics. This can then result in modifications from the contrastive set's interventions in the test data to be more straightforwardly mapped and identified by the TLDR's NLI module.

For a more detailed error analysis and exploration of each model's performance across various sections of the clinical trial, refer to Appendix~\ref{sec:appendix}. The main takeaway is that TLDR methods leveraging fine-tuning and distinct premise summarization, consistently outperformed simpler input models across all clinical trial sections, demonstrating the significance of specialized summarization and training techniques in managing the challenges of lengthy premises in clinical trials.

\section{Conclusion}
% \vspace{-0.6em}
In this paper, we introduced TLDR (\textbf{T}5-generated clinical-\textbf{L}anguage summaries for \textbf{D}eBERTa \textbf{R}eport Analysis) tailored for clinical NLI tasks, with a focus on NLI4CT 2024. Our investigation reveals that strategies incorporating SciFive for distinct premise summarization and fine-tuning for summarization to better align with entailed statements about the premises markedly improve handling of clinical language and reasoning complexities. Despite the challenges posed by lengthy premises in clinical reports, our TLDR methods, particularly those employing advanced summarization through fine-tuning, consistently demonstrated superior performance over simpler methods. This underscores the importance of model adaptation and the strategic selection of summarization techniques in enhancing the accuracy and reliability of NLI tasks within the clinical domain.

Looking ahead, a promising avenue for future work involves the use of the best encoder-decoder transformer summarization model for each specific section of clinical reports. This approach could potentially improve the overall performance of NLI where we saw fine-tuned SciFive was particularly better in some sections of the clinical report and T5-based summaries were better at some other sections. Continued exploration and refinement of these models are essential to further advance the field of NLI in clinical applications.

\section{Acknowledgements}
The authors would also thank Dr. Daniel Fried, Saujas Vaduguru, and Kundan Krishna for helpful discussions.

S.N. was supported by Natural Sciences and Engineering Research Council of Canada .

\bibliography{anthology, custom}

% \clearpage

\appendix

\section{Fine-Tuning Encoder-Decoder Transformers for Specific Summary Generation}
In this section, we present how the NLI performance varies across different fine-tuning steps of the T5 variants.

In the fine-tuning results of Table~\ref{tab:finetuning}, we observe a peak performance at two epochs for both T5-small and SciFive-base models, with a slight performance drop as the number of epochs increases. The T5-small model shows a trade-off between precision and recall, reaching its highest recall at five epochs but with better overall performance at two epochs. The SciFive-base model maintains a consistent precision after two epochs, but the recall fluctuates, suggesting that the optimal number of training epochs is crucial for maintaining model balance and preventing overfitting.
We believe the reason that SciFive required longer fine-tuning epochs compared to T5-small is because the initial fine-tuning of SciFive diminished its summarization capabilities relative to T5-small, necessitating extended fine-tuning to restore performance.

\begin{table}[!ht]
\centering
\caption{Fine-Tuning Results}
\label{tab:finetuning}
\adjustbox{max width=1\linewidth}{
\begin{tabular}{|l||c|c|c|c|c|}
\hline
\textbf{Method} & \textbf{Macro F1} & \textbf{Precision} & \textbf{Recall} & \textbf{Faithfulness} & \textbf{Consistency} \\ \hline \hline
\multicolumn{6}{|c|}{\textbf{T5 Fine-Tuning}} \\ \hline
TLDR (T5-small - 2 Epochs) &  0.605 & 0.580 & 0.633 & 0.557 & 0.593 \\ \hline
TLDR (T5-small - 5 Epochs) &  0.635 & 0.676 & 0.599 & 0.436 & 0.557 \\ \hline
TLDR (T5-small - 7 Epochs) &  0.601 & 0.564 & 0.644 & 0.597 & 0.590 \\ \hline
TLDR (T5-small - 10 Epochs) &  0.603 & 0.580 & 0.628 & 0.608 & 0.587 \\ \hline
\multicolumn{6}{|c|}{\textbf{SciFive Fine-Tuning}} \\ \hline
TLDR (SciFive-base - 2 Epochs) &  0.570 & 0.528 & 0.620 & 0.552 & 0.580 \\ \hline
TLDR (SciFive-base - 5 Epochs) &  0.628 & 0.652 & 0.606 & 0.470 & 0.563 \\ \hline
TLDR (SciFive-base - 7 Epochs) &  0.613 & 0.588 & 0.639 & 0.560 & 0.589 \\ \hline
TLDR (SciFive-base - 10 Epochs) &  0.658 & 0.684 & 0.633 & 0.501 & 0.581 \\ \hline
\end{tabular}}
\end{table}

\section{Error Analysis}
\label{sec:appendix}
In this discussion, we first delve into a detailed error analysis in subsection~\ref{sec:error_analysis}, examining model performances across different sections and types of clinical trial data. The results presented is on the practice held-out test set. Following this, in subsection~\ref{sec:model_agreement} we explore the prediction agreement among the various TLDR models.

\subsection{Error Analysis}
\label{sec:error_analysis}
The first part of our error analysis focuses on the performance of TLDR methods and ablation DeBERTa models across different sections of clinical trial reports: Eligibility, Adverse Events, Results, and Interventions. The analysis is grounded in Macro F1 scores, average premise lengths, and average statement lengths, as detailed in the Tables~\ref{tab:section_eligibility},\ref{tab:section_adverse_events},\ref{tab:section_results},\ref{tab:section-intervention}.

\paragraph{Eligibility Section}
In the Eligibility section, the TLDR method using flan-T5-base for zero-shot distinct premises achieved the highest Macro F1 score (0.626), indicating its effectiveness in summarizing and understanding eligibility criteria. Notably, this model and the best fine-tuned SciFive-base model managed to significantly reduce the average premise length while maintaining high performance, suggesting that effective summarization can aid in dealing with long premises typically found in this section.

\paragraph{Adverse Events Section}
The Adverse Events section saw the highest Macro F1 score (0.775) with the TLDR method using T5-small, best fine-tuned. This model also had one of the shortest average premise lengths, indicating that fine-tuning on specific data, even with shorter premise lengths, can yield high accuracy in identifying adverse events. The DeBERTa models performed relatively well in this section but were outperformed by the TLDR approaches.

\paragraph{Results Section}
For the Results section, the TLDR method with fine-tuned SciFive-base, showed the best performance with a Macro F1 of 0.67. Interestingly, this model also had the shortest average premise length, suggesting a strong correlation between effective summarization and model performance. The low performance of the SciFive-base zero-shot model indicates that domain-specific fine-tuning is crucial for understanding complex results data that we gain from generating summaries similar to the entailment statements. Note that similar to \cite{zhou-etal-2023-thifly} where they observed SciFive showed superior performance for results with numerical data, we also see the gain in using SciFive when used for the results section.

\paragraph{Intervention Section}
In the Intervention section, the TLDR flan-T5-base method for zero-shot combined premises showed the highest Macro F1 score (0.647). This suggests that the model's ability to synthesize information from combined premises is particularly effective in understanding intervention-related data.

In the second part of our error analysis, distinct trends are revealed in the performance of the DeBERTa and TLDR models when dealing with single and comparison premises in clinical trial reports. This distinction is crucial as single premises present a straightforward context, whereas comparison premises involve juxtaposing and interpreting two separate contexts. These results are presented in the appendix in Table~\ref{tab:type_comparison}.

\paragraph{Single Premises}
In the single premise category, the TLDR methods generally outperform their ablated counterparts using DeBERTa models. The TLDR method using flan-T5-base for zero-shot distinct premises and the best fine-tuned SciFive-base model both achieved a Macro F1 score of 0.642, the highest in this category. This indicates their robustness in handling singular, focused clinical contexts. Notably, these models significantly reduced the average premise length, with the best fine-tuned SciFive-base model achieving the shortest length, which suggests an effective summarization capability that preserves essential information that we achieve by attempting to generate summaries that are aligned with the entailment statements.

\paragraph{Comparison Premises}
For comparison premises, where the task involves analyzing and relating two different contexts, the TLDR models still outperform the DeBERTa models, but with a slight decrease in overall effectiveness compared to single premises. The highest Macro F1 score is 0.631 with the TLDR flan-T5-base for zero-shot distinct premises. The best fine-tuned models, both T5-small and SciFive-base, also show strong performance in this more complex scenario. Interestingly, the average premise lengths are longer for comparison premises across all models, underscoring the increased complexity and information content in these types of premises.

Across both single and comparison premises, TLDR methods demonstrate superior performance, especially in handling and effectively summarizing complex clinical data. The shorter average premise lengths in the best-performing models suggest that their summarization strategies are successful in distilling essential information without losing context crucial for NLI tasks. This is particularly evident in the comparison premises, where managing two contexts simultaneously is a challenging task. In conclusion, the type of premise (single or comparison) significantly impacts the model's performance, with TLDR methods showing robustness in both scenarios. The findings emphasize the importance of tailored summarization techniques and model fine-tuning to handle the varying complexities in clinical trial reports.

\subsection{Results Split By Type}
Below in Table~\ref{tab:type_comparison}  we include results that were obtained split on whether the data instance was a single instance meaning only a primary premise was given or a comparison instance where both a primary and a secondary instance was given. 
\subsection{Results Split By Section Type}
Below in Tables~\ref{tab:section_eligibility},~\ref{tab:section_adverse_events}~\ref{tab:section_results},~\ref{tab:section-intervention}  we include results that were obtained split on Eligibility, Adverse Events, Results, and Intervention sections respectively. These are the different sections in the Clinical Trial Reports that the statement in the data instance is referring to

\begin{table*}
\centering
\caption{Type Results}
\label{tab:type_comparison}
\adjustbox{max width=1\linewidth}{
\begin{tabular}{|l||c|c|c|c|c|}
\hline
\textbf{Method} & \textbf{Macro F1} & \textbf{Avg Premise Len} & \textbf{Avg Premise - Ent} & \textbf{Avt Premise - Con} & \textbf{Avg Statement Len} \\ \hline \hline
\multicolumn{6}{|l|}{\textbf{Single}} \\ \hline
\multicolumn{6}{|c|}{\textbf{Ablations: DeBERTa Methods}} \\ \hline
DeBERTa + Truncated Premise(s) & 0.484 & 1152.2 & 1149.5 & 1154.9 & 121.2 \\ \hline
DeBERTa + Extractive Summarized Premise(s) & 0.574 & 725.0 & 724.1 & 725.8 & 121.2 \\ \hline
\multicolumn{6}{|c|}{\textbf{TLDR Methods}} \\ \hline
TLDR (flan-T5-base - Zero-Shot Combined Premises) & 0.642 & 334.0 & 333.3 & 334.8 & 121.2 \\ \hline
TLDR (flan-T5-base - Zero-Shot Distinct Premises) & 0.644 & 276.2 & 275.8 & 276.6 & 121.2 \\ \hline
TLDR (SciFive-base - Zero-Shot Premises) & 0.427 & 542.6 & 545.3 & 539.9 & 121.2 \\ \hline
TLDR (T5-small - Best fine-tuned) & 0.637 & 196.0 & 196.2 & 195.9 & 121.2  \\ \hline
TLDR (SciFive-base - Best fine-tuned) & 0.642 & 79.2 & 79.1 & 79.3 & 121.2 \\ \hline \hline
\multicolumn{6}{|l|}{\textbf{Comparison}} \\ \hline
\multicolumn{6}{|c|}{\textbf{DeBERTa Methods}} \\ \hline
DeBERTa + Truncated Premise(s) & 0.52 & 2270.8 & 2273.9 & 2267.8 & 145.7 \\ \hline
DeBERTa + Extractive Summarized Premise(s) & 0.522 & 956.5 & 958.2 & 954.9 & 145.7 \\ \hline
\multicolumn{6}{|c|}{\textbf{TLDR Methods}} \\ \hline
TLDR (flan-T5-base - Zero-Shot Combined Premises) & 0.587 & 407.5 & 406.4 & 408.7 & 145.7 \\ \hline
TLDR (flan-T5-base - Zero-Shot Distinct Premises) & 0.631 & 448.9 & 449.1 & 448.8 & 145.7 \\ \hline
TLDR (SciFive-base - Zero-Shot Premises) & 0.471 & 1046.1 & 1043.5 & 1048.8 & 145.7 \\ \hline
TLDR (T5-small - Best fine-tuned) & 0.626 & 371.7 & 371.5 & 372.0 & 145.7  \\ \hline
TLDR (SciFive-base - Best fine-tuned) & 0.612 & 150.5 & 150.2 & 150.7 & 145.7 \\ \hline

\end{tabular}}
\end{table*}

\begin{table*}
\centering
\caption{Section Results - Eligibility}
\label{tab:section_eligibility}
\adjustbox{max width=1\linewidth}{
\begin{tabular}{|l||c|c|c|c|c|}
\hline
\multicolumn{6}{|c|}{\textbf{Eligibility}} \\ \hline
\textbf{Method} & \textbf{Macro F1} & \textbf{Avg Premise Len} & \textbf{Avg Premise - Ent} & \textbf{Avt Premise - Con} & \textbf{Avg Statement Len} \\ \hline \hline
\multicolumn{6}{|c|}{\textbf{Ablations: DeBERTa Methods}} \\ \hline
DeBERTa + Truncated Premise(s) & 0.395 & 3776.0 & 3776.0 & 3776.0 & 147.4 \\ \hline
DeBERTa + Extractive Summarized Premise(s) & 0.444 & 1517.7 & 1517.7 & 1517.7 & 147.4 \\ \hline
\multicolumn{6}{|c|}{\textbf{TLDR Methods}} \\ \hline
TLDR (flan-T5-base - Zero-Shot Combined Premises) & 0.574 & 636.6 & 636.6 & 636.6 & 147.4 \\ \hline
TLDR (flan-T5-base - Zero-Shot Distinct Premises) &  0.626 & 418.8 & 418.8 & 418.8 & 147.4 \\ \hline
TLDR (SciFive-base - Zero-Shot Premises) & 0.448 & 1070.3 & 1070.3 & 1070.3 & 147.4 \\ \hline
TLDR (T5-small - Best fine-tuned) &  0.537 & 383.0 & 383.0 & 383.0 & 147.4  \\ \hline
TLDR (SciFive-base - Best fine-tuned) & 0.613 & 137.3 & 137.3 & 137.3 & 147.4 \\ \hline
\end{tabular}}
\end{table*}

\begin{table*}
\centering
\caption{Section Results - Adverse Events}
\label{tab:section_adverse_events}
\adjustbox{max width=1\linewidth}{
\begin{tabular}{|l||c|c|c|c|c|}
\hline
\multicolumn{6}{|c|}{\textbf{Adverse Events}} \\ \hline
\textbf{Method} & \textbf{Macro F1} & \textbf{Avg Premise Len} & \textbf{Avg Premise - Ent} & \textbf{Avt Premise - Con} & \textbf{Avg Statement Len} \\ \hline \hline
\multicolumn{6}{|c|}{\textbf{Ablations: DeBERTa Methods}} \\ \hline
DeBERTa + Truncated Premise(s) & 0.583 & 496.1 & 496.1 & 496.1 & 109.9 \\ \hline
DeBERTa + Extractive Summarized Premise(s) & 0.646 & 496.6 & 496.6 & 496.6 & 109.9 \\ \hline
\multicolumn{6}{|c|}{\textbf{TLDR Methods}} \\ \hline
TLDR (flan-T5-base - Zero-Shot Combined Premises) & 0.641 & 243.0 & 243.0 & 243.0 & 109.9 \\ \hline
TLDR (flan-T5-base - Zero-Shot Distinct Premises) & 0.699 & 292.4 & 292.4 & 292.4 & 109.9 \\ \hline
TLDR (SciFive-base - Zero-Shot Premises) & 0.43 & 678.1 & 678.1 & 678.1 & 109.9 \\ \hline
TLDR (T5-small - Best fine-tuned) &  0.775 & 217.2 & 217.2 & 217.2 & 109.9  \\ \hline
TLDR (SciFive-base - Best fine-tuned) & 0.675 & 107.0 & 107.0 & 107.0 & 109.9 \\ \hline
\end{tabular}}
\end{table*}

\begin{table*}
\centering
\caption{Section Results - Results}
\label{tab:section_results}
\adjustbox{max width=1\linewidth}{
\begin{tabular}{|l||c|c|c|c|c|}
\hline
\multicolumn{6}{|c|}{\textbf{Results}} \\ \hline
\textbf{Method} & \textbf{Macro F1} & \textbf{Avg Premise Len} & \textbf{Avg Premise - Ent} & \textbf{Avt Premise - Con} & \textbf{Avg Statement Len} \\ \hline \hline
\multicolumn{6}{|c|}{\textbf{Ablations: DeBERTa Methods}} \\ \hline
DeBERTa + Truncated Premise(s) & 0.45 & 2022.4 & 2013.9 & 2030.8 & 145.5 \\ \hline
DeBERTa + Extractive Summarized Premise(s) & 0.518 & 971.4 & 971.6 & 971.1 & 145.5 \\ \hline
\multicolumn{6}{|c|}{\textbf{TLDR Methods}} \\ \hline
TLDR (flan-T5-base - Zero-Shot Combined Premises) & 0.575 & 358.0 & 352.6 & 363.3 & 145.5 \\ \hline
TLDR (flan-T5-base - Zero-Shot Distinct Premises) & 0.622 & 437.7 & 435.6 & 439.8 & 145.5 \\ \hline
TLDR (SciFive-base - Zero-Shot Premises) & 0.333 & 1040.8 & 1035.0 & 1046.6 & 145.5 \\ \hline
TLDR (T5-small - Best fine-tuned) &  0.604 & 320.2 & 318.2 & 322.2 & 145.5  \\ \hline
TLDR (SciFive-base - Best fine-tuned) & 0.67 & 132.0 & 130.5 & 133.5 & 145.5 \\ \hline
\end{tabular}}
\end{table*}

\begin{table*}
\centering
\caption{Section Results - Intervention}
\label{tab:section-intervention}
\adjustbox{max width=1\linewidth}{
\begin{tabular}{|l||c|c|c|c|c|}
\hline
\multicolumn{6}{|c|}{\textbf{Intervention}} \\ \hline
\textbf{Method} & \textbf{Macro F1} & \textbf{Avg Premise Len} & \textbf{Avg Premise - Ent} & \textbf{Avt Premise - Con} & \textbf{Avg Statement Len} \\ \hline \hline
\multicolumn{6}{|c|}{\textbf{Ablations: DeBERTa Methods}} \\ \hline
DeBERTa + Truncated Premise(s) & 0.558 & 752.9 & 752.9 & 752.9 & 135.1 \\ \hline
DeBERTa + Extractive Summarized Premise(s) & 0.543 & 439.1 & 439.1 & 439.1 & 135.1 \\ \hline
\multicolumn{6}{|c|}{\textbf{TLDR Methods}} \\ \hline
TLDR (flan-T5-base - Zero-Shot Combined Premises) & 0.647 & 252.1 & 252.1 & 252.1 & 135.1\\ \hline
TLDR (flan-T5-base - Zero-Shot Distinct Premises) & 0.602 & 339.0 & 339.0 & 339.0 & 135.1 \\ \hline
TLDR (SciFive-base - Zero-Shot Premises) & 0.516 & 526.7 & 526.7 & 526.7 & 135.1 \\ \hline
TLDR (T5-small - Best fine-tuned) & 0.615 & 246.9 & 246.9 & 246.9 & 135.1  \\ \hline
TLDR (SciFive-base - Best fine-tuned) & 0.555 & 98.3 & 98.3 & 98.3 & 135.1 \\ \hline
\end{tabular}}
\end{table*}

\subsection{Prediction agreement across various model}
\label{sec:model_agreement}

\begin{figure}[ht!]
\centering
\includegraphics[width=0.85\linewidth]{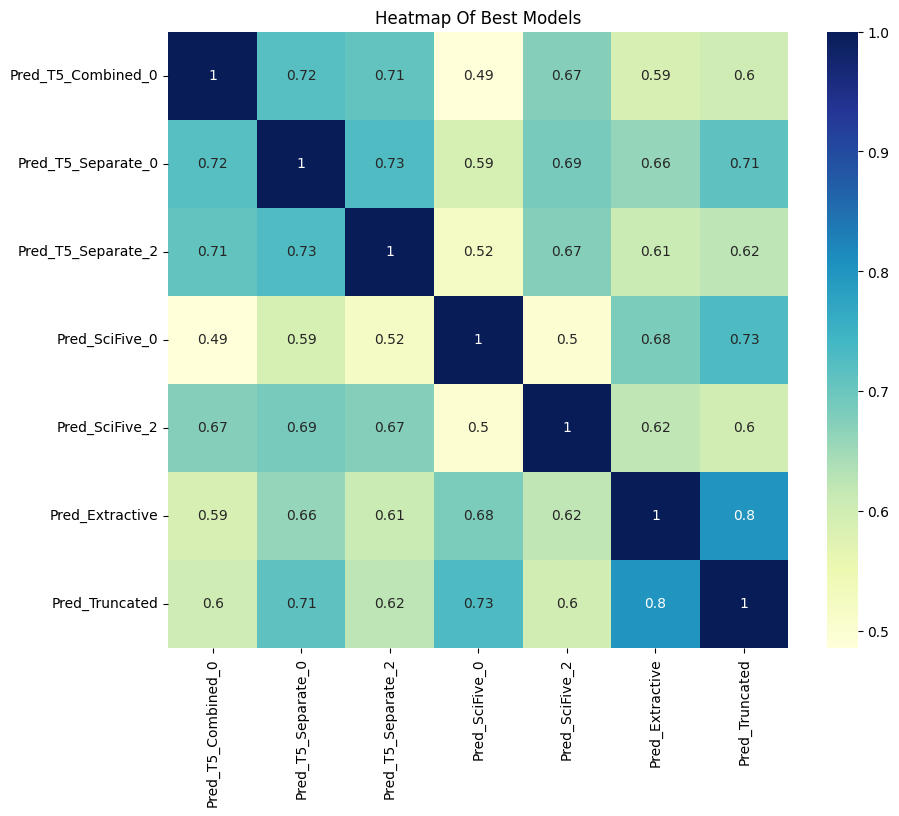}
\caption{Heatmap comparing the model predictions}
\label{fig:heatmap}
\end{figure}

In our comparative analysis of model predictions, depicted in Figure~\ref{fig:heatmap}, we observe distinct patterns of agreement among the various models tested. Notably, T5-based models exhibit a high degree of consistency in their predictions, as evidenced by the darker blue 3x3 square in the top left corner of the heatmap. This suggests a strong underlying similarity in how these models process and interpret the summaries for the test data. In contrast, the SciFive-based models display a marked divergence in their prediction patterns. The fine-tuned version of the SciFive model, in particular, demonstrates a significant shift in its predictions, aligning with the positive performance changes highlighted in previous sections. Furthermore, the two ablated versions employing either truncated premises or extractive summarization exhibit a high level of agreement in their predictions, as indicated by the dark blue 2x2 square in the heatmap's bottom right corner. This consistency points to the robustness of these ablation methods in maintaining prediction alignment. Overall, these findings underscore the varying degrees of prediction agreement across different model architectures and highlight the impact of model-specific features and training approaches on prediction outcomes in clinical NLI tasks.

\end{document}